\documentclass[journal]{IEEEtran}
\usepackage{cite}
\usepackage{amsmath,amssymb,amsfonts}
\usepackage{graphicx}
\usepackage{textcomp}

\usepackage{booktabs}
\usepackage{tabularx}
\usepackage{multirow}
\usepackage[normalem]{ulem}
\usepackage{makecell}
\usepackage{algorithm}
\usepackage{algpseudocode}
\usepackage{url}
\usepackage{pifont}

\usepackage{hyperref}

\begin{document}

\title{PRIME: Prototype-Driven Multimodal Pretraining for Cancer Prognosis with Missing Modalities}

\author{Kai Yu, Shuang Zhou, Yiran Song, Zaifu Zhan, Jie Peng, Kaixiong Zhou, Tianlong Chen, Feng Xie, Meng Wang, Huazhu Fu, \IEEEmembership{Senior Member, IEEE}, Mingquan Lin, Rui Zhang
\thanks{This work was supported by the National Institutes of Health’s National Center for Complementary and Integrative Health under grant number R01AT009457, National Institute on Aging under grant number R01AG078154, and National Cancer Institute under grant number R01CA287413.}
\thanks{Kai Yu, Shuang Zhou, Yiran Song, Zaifu Zhan, Feng Xie, Mingquan Lin, and Rui Zhang are with the Division of Computational Health Sciences, Department of Surgery, University of Minnesota, Minneapolis, MN, USA. Jie Peng and Tianlong Chen are with the University of North Carolina at Chapel Hill, Chapel Hill, NC, USA. Kaixiong Zhou is with North Carolina State University, Raleigh, NC, USA. Meng Wang is with the Centre for Innovation \& Precision Eye Health, Department of Ophthalmology, Yong Loo Lin School of Medicine, National University of Singapore, Singapore. Huazhu Fu is with the Institute of High Performance Computing, Agency for Science, Technology and Research, Singapore.}
\thanks{Mingquan Lin and Rui Zhang are the co-corresponding authors (e-mail: lin01231@umn.edu, ruizhang@umn.edu)}}
\maketitle

\begin{abstract}
Multimodal self-supervised pretraining offers a promising route to cancer prognosis by integrating histopathology whole-slide images, gene expression, and pathology reports, yet most existing approaches require fully paired and complete inputs. 
In practice, clinical cohorts are fragmented and often miss one or more modalities, limiting both supervised fusion and scalable multimodal pretraining. 
We propose PRIME, a missing-aware multimodal self-supervised pretraining framework that learns robust and transferable representations from partially observed cohorts. 
PRIME maps heterogeneous modality embeddings into a unified token space and introduces a shared prototype memory bank for latent-space semantic imputation via patient-level consensus retrieval, producing structurally aligned tokens without reconstructing raw signals. 
Two complementary pretraining objectives: inter-modality alignment and post-fusion consistency under structured missingness augmentation, jointly learn representations that remain predictive under arbitrary modality subsets. 
We evaluate PRIME on The Cancer Genome Atlas with label-free pretraining on 32 cancer types and downstream 5-fold evaluation on five cohorts across overall survival prediction, 3-year mortality classification, and 3-year recurrence classification. 
PRIME achieves the best macro-average performance among all compared methods, reaching 0.653 C-index, 0.689 AUROC, and 0.637 AUROC on the three tasks, respectively, while improving robustness under test-time missingness and supporting parameter-efficient and label-efficient adaptation. 
These results support missing-aware multimodal pretraining as a practical strategy for prognosis modeling in fragmented clinical data settings.

\end{abstract}


\section{Introduction}
\label{sec:introduction}
\IEEEPARstart{A}{ccurate} cancer prognosis is central to personalized treatment planning and follow-up scheduling~\cite{Zhang2023-du}. 
In clinical practice, prognosis is assessed through multiple endpoints, including time-to-event outcomes such as overall survival (OS) and progression-free interval (PFI), as well as fixed-horizon outcomes that are often more directly actionable (e.g., 3-year mortality and recurrence)~\cite{Harrell1996-qi}. 
Developing reliable models for these endpoints remains challenging because tumors are highly heterogeneous, labeled outcomes are limited, and predictive signals must often be extracted from heterogeneous and complementary data sources~\cite{Boehm2022-cq}.

Large-scale resources such as The Cancer Genome Atlas (TCGA) provide an opportunity to learn prognostic signals from complementary modalities~\cite{Tomczak2015-ix}. 
Histopathology whole-slide images (WSIs) capture spatial morphology and tumor microenvironment patterns, bulk RNA sequencing reflects molecular programs, and pathology reports summarize diagnostic findings in natural language. 
Jointly leveraging these modalities has repeatedly shown advantages over unimodal modeling~\cite{Xu2025-ib,Zhou2023-eb,Raza2025-ki}.
However, real-world multimodal cohorts are inherently incomplete: modalities are often missing due to cost, assay availability, retrospective collection, or incomplete documentation. Consequently, the intersection of fully paired samples can be substantially smaller than the union of patients with at least one modality, limiting both supervised fusion and scalable representation learning~\cite{Ruffini2026-tf,Qu2026-lf}. This fragmentation poses a particular challenge for multimodal pretraining, where cross-modal objectives typically assume complete pairing.

Most existing multimodal prognosis pipelines focus on supervised fusion at the downstream stage, combining modality-specific features via early/late fusion or attention-based intermediate fusion~\cite{Nikolaou2025-ss,Steyaert2023-dy}. 
These approaches handle missingness using heuristic placeholders, leading to unstable performance under incomplete observations~\cite{Qu2026-cr}. 
A natural remedy is to improve representations through self-supervised pretraining. However, extending pretraining to multimodal settings faces a key obstacle: common cross-modal objectives require paired observations, restricting learning to the intersection of modalities and preventing the model from exploiting partially observed cohorts~\cite{Ruffini2026-tf,Xu2025-ib}. 
Moreover, most strategies address missing-modality robustness only implicitly, rather than explicitly optimizing representations that remain consistent and predictive under arbitrary modality subsets~\cite{Vale-Silva2021-er,Ruffini2026-tf}.

In this work, we treat missingness not as an exception but as a structural property of clinical data, and propose PRIME, a missing-aware multimodal self-supervised pretraining framework for WSI-RNA-report learning. 
PRIME operates in a unified token space and introduces a shared prototype memory bank as a latent interface across heterogeneous modalities, where each prototype is a learnable token sequence. 
For a patient with incomplete modalities, PRIME retrieves and aggregates information from a shared prototype bank based on observed evidence, synthesizing plausible latent tokens for missing modalities without reconstructing raw signals. 
Unlike generative imputation, which requires modality-specific decoders and risks hallucination, prototype-based imputation operates entirely in the latent space and naturally preserves structural alignment across modalities.

Building on this mechanism, PRIME optimizes missing-aware pretraining with two complementary components: 
(i) inter-modality alignment computed on paired modality subsets to scale learning with partially observed cohorts while avoiding imputation noise, and 
(ii) post-fusion consistency under structured missingness augmentation, where modality- and token-level dropout create two views and dropped elements are imputed using Dirichlet-driven prototype mixtures to obtain diverse yet semantically anchored representations. 
Together, these designs learn robust and transferable multimodal representations that support inference under arbitrary missing-modality subsets and enable parameter-efficient adaptation (e.g., linear probing) in label-limited settings.

We evaluate PRIME on TCGA across three clinically relevant endpoints: (i) OS time-to-event prediction evaluated by concordance index (C-index), (ii) 3-year mortality classification evaluated by AUROC, and (iii) 3-year recurrence classification based on PFI annotations evaluated by AUROC. 
Beyond full-modality testing, we perform robustness experiments by removing modalities at inference time, and we further assess label efficient transfer and parameter efficiency. 
Across tasks, PRIME consistently improves downstream performance after self-supervised pretraining, mitigates performance degradation under missing modalities, and enables parameter-efficient transfer in which linear probing remains competitive with, and can surpass, full fine-tuning.

Our contributions are as follows: 
 \begin{itemize}
     \item We propose PRIME, a missing-aware multimodal self-supervised pretraining framework that leverages partially observed cohorts for cancer prognosis and supports arbitrary modality subsets at inference.
     \item We introduce a learnable prototype memory bank with patient-level consensus retrieval for latent-space semantic imputation, and design two complementary pretraining objectives: inter-modality alignment and post-fusion consistency, to learn robust and transferable multimodal representations under missingness.
     \item We evaluate PRIME on TCGA across three clinically relevant endpoints, where it achieves the best macro-average performance among all compared methods while demonstrating improved robustness under test-time missingness and supporting label-efficient and parameter-efficient adaptation.
 \end{itemize}

\section{Related Work}
\subsection{Multimodal Prognostic Modeling}
Multimodal cancer prognosis modeling has been widely studied on cohorts such as TCGA by integrating WSIs, molecular profiles, and pathology text. A broad spectrum of fusion strategies has been explored, from early/late fusion and tensor-based or attention-based interaction models~\cite{Lahat2015-uo,Lipkova2022-so,Zadeh2017-dq,Rahman2020-sm,Tsai2019-py}, to task-specific pathology-omics architectures such as MCAT~\cite{Chen2021-rx}, Porpoise~\cite{Chen2022-pw}, and PathOmics~\cite{Ding2023-ij}.
More recently, many pipelines adopt frozen unimodal encoders to extract modality-specific embeddings and train lightweight fusion modules, which is computationally efficient and aligns with practical deployment constraints~\cite{Qu2026-cr,Xu2025-ib,SongUnknown-db}.

However, these approaches are architecturally designed for complete inputs: fusion modules expect a fixed set of modality features, and missingness is addressed only as a post hoc workaround, typically by taking zeros or mean features for absent modalities~\cite{Qu2026-cr}. Such heuristic placeholders inject uninformative signals into the fusion process, degrading robustness and limiting effective use of partially observed patients.

\subsection{Multimodal Foundation Models and Pretraining}
Computational pathology has witnessed rapid development of foundation models trained with large-scale self-supervision. Vision-only and vision-language pretraining has produced transferable WSI encoders (e.g., UNI, CONCH)~\cite{Chen2024-ev,Lu2024-vx}, and multimodal pretraining further seeks to align WSIs with molecular and clinical-text modalities for downstream clinical prediction~\cite{Xu2025-ib,Zhou2025-co,Wang2025-hw}.
A representative tri-modal framework, mSTAR~\cite{Xu2025-ib}, integrates WSIs, gene-expression features, and pathology reports via cross-modal alignment and self-taught distillation; however, its alignment stage relies on paired modalities, excluding patients who lack a complementary modality, and missing-modality benefits are realized through knowledge injection into a single-modality encoder rather than optimizing fused representations across arbitrary modality subsets.
Pathology-omics frameworks such as POMP~\cite{Wang2025-hw} combine cross-modal contrastive alignment with masking-based modeling to improve robustness to partial corruption, while MICE~\cite{Zhou2025-co} explores mixture-of-experts for prognostic representation learning. Despite these advances, scalable pretraining that explicitly exploits the union of partially observed cohorts while ensuring consistent inference under arbitrary modality subsets remains an open challenge.

\begin{figure*}[!t]
\centerline{\includegraphics[width=\textwidth]{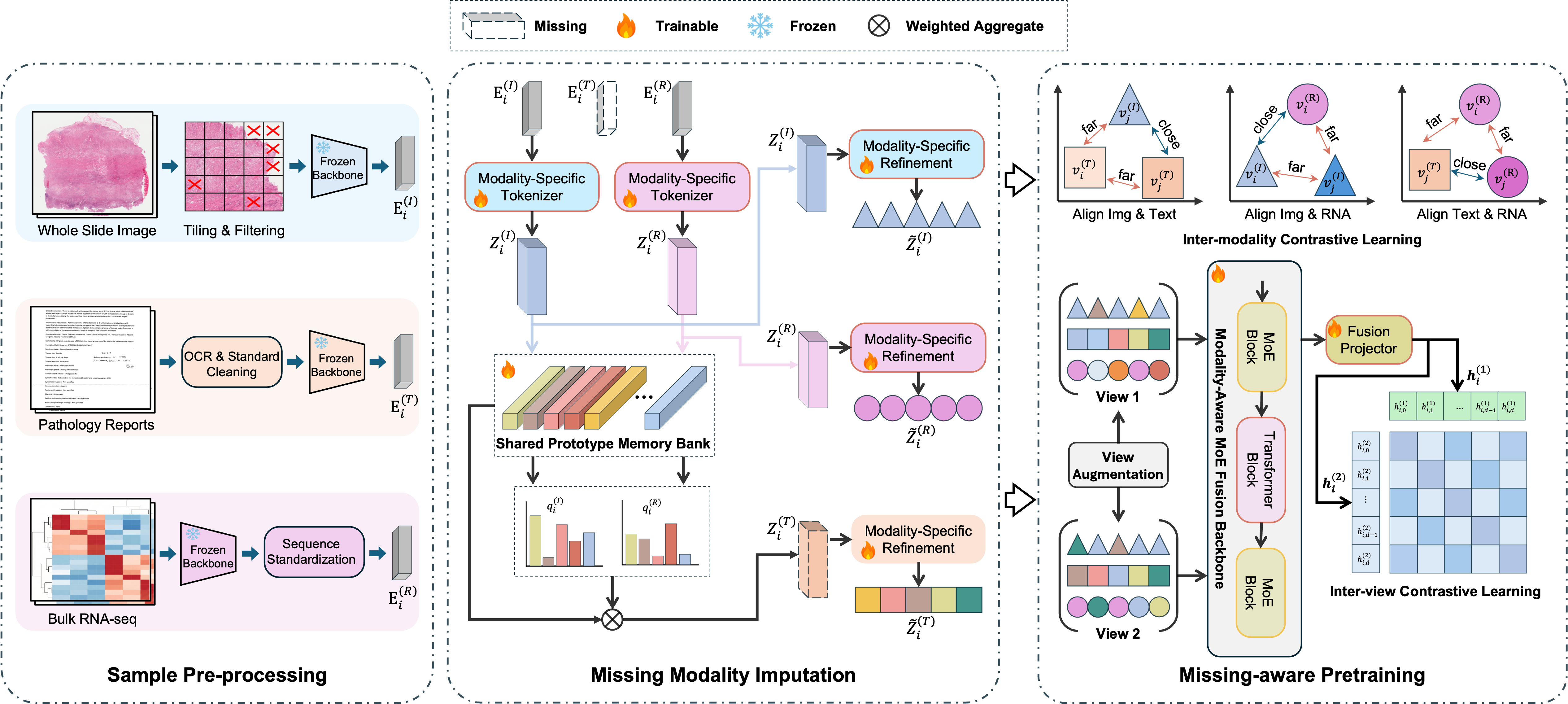}}
\caption{Overview of the proposed PRIME framework for multimodal cancer prognosis with missing modalities. \textbf{Left (Sample Pre-processing):} Frozen encoders extract token embeddings from WSIs, RNA-seq, and pathology reports. \textbf{Middle (Missing Modality Imputation):} In a unified token space, a shared sequence-level prototype memory bank imputes missing modalities via patient-level consensus retrieval, followed by modality-specific refinement. \textbf{Right (Missing-aware Pretraining):} Two complementary 
objectives, inter-modality contrastive alignment and post-fusion consistency under modality/token dropout with prototype mixtures, jointly drive self-supervised pretraining.}
\label{fig1}
\end{figure*}

\subsection{Learning with Missing Modalities}
Missing-modality learning has been addressed through several strategies. Generative methods synthesize absent modalities from observed ones but risk reconstruction-prediction mismatch and error propagation in high-dimensional settings~\cite{Zhou2025-dz}. Retrieval- or memory-based approaches leverage reference patients to approximate missing features, though performance depends on reference-set coverage and similarity metric stability~\cite{Qu2026-lf}. Modality-aware prompts or pseudo-embeddings adapt models to variable availability during downstream training but do not address how to scale multimodal pretraining on fragmented cohorts~\cite{Qu2026-cr}. Masked modeling and modality dropout improve robustness by exposing models to systematic incompleteness~\cite{Liu2023-iy, Robinet2025-tz}, yet adapting these designs to WSI-omics-text learning introduces additional challenges including large semantic gaps, heterogeneous token structures, and incomplete cross-modal pairing. In particular, a unified framework that performs missing-aware pretraining on partially observed cohorts while explicitly optimizing fused representations to remain consistent under arbitrary modality subsets remains limited.

\section{Method}
\subsection{Overview}
As illustrated in Fig.~\ref{fig1}, PRIME considers three 
modalities: histopathology WSIs (Image), bulk RNA sequencing (RNA), and pathology reports (Text). Frozen encoders extract modality-specific embeddings, a shared prototype memory bank imputes missing modalities in latent space, and two contrastive objectives drive missing-aware pretraining. The pretrained backbone supports arbitrary modality subsets via full fine-tuning or linear probing.

\subsection{Input Embeddings and Notation}
For each patient $i$, let the complete set of multimodality inputs be denoted as $\mathcal{M}=\{I,R,T\}$, corresponding to image, RNA, and text, respectively. To facilitate the subsequent multimodal fusion and imputation, we precompute modality-specific feature embeddings
using frozen backbone encoders and use these embeddings as inputs.
For modality $m\in\mathcal{M}$, the input is a fixed-length token sequence
\begin{equation}
\mathbf{E}_i^{(m)} \in \mathbb{R}^{L_m \times D_m},
\end{equation}
where $L_m$ is the token length and $D_m$ is the feature dimension.
If a modality is unavailable for patient $i$, we set $\mathbf{E}_i^{(m)}=\varnothing$ and record its
availability by a binary mask $a_{i,m} \in \{0, 1\}$.
Details of feature extraction and preprocessing are provided in Sec.~IV-A.

\subsection{Missing Modality Imputation}
In real-world clinical settings, patients often lack certain data modalities. To fully utilize incomplete records, we propose a dynamic imputation mechanism via a shared prototype memory bank. 
For patient $i$, let $\mathcal{S}_i \subseteq \mathcal{M}$ denote the available modalities. 

\subsubsection{Modality-Specific Tokenization}
Because the pre-computed embeddings $\mathbf{E}_i^{(m)}$ have varying sequence lengths, we employ a modality-specific query-based cross-attention tokenizer to map each available modality $m \in \mathcal{S}_i$ into a unified semantic space. 
Each tokenizer uses $T_q$ learnable query tokens 
$\mathbf{Q}_m \in \mathbb{R}^{T_q \times D}$ (with $T_q=128$ and hidden dimension $D=512$) as queries in a cross-attention layer followed by a feed-forward network, condensing the variable-length input into a fixed-length representation:
\begin{equation}
\mathbf{Z}_i^{(m)} = \mathrm{FFN}(\mathrm{MHCA}
(\mathrm{Q}=\mathbf{Q}_{m}, \mathrm{K,V}=\mathbf{E}_i^{(m)})) 
\in \mathbb{R}^{T_q \times D}
\end{equation}
This process structurally aligns all modalities by extracting exactly $T_q$ salient tokens per input.

\subsubsection{Cross-Modal Consensus via Shared Prototype Bank}
To bridge the semantic gap between different modalities and guide the imputation of missing data, we introduce a learnable shared prototype memory bank (a sequence-level codebook) $\mathcal{C} = \{\mathbf{C}_k\}_{k=1}^{K_c}$, where each prototype is a token sequence $\mathbf{C}_k \in \mathbb{R}^{T_q \times D}$, and $K_c$ is the total number of prototypes.

For an observed modality with token sequence $\mathbf{Z}_i^{(m)}$, we compute a soft assignment over the prototypes based on their mean-pooled token features:
\begin{equation}
\bar{\mathbf{Z}}_i^{(m)} = \mathrm{norm}\!\left(\frac{1}{T_q}\sum_{t=1}^{T_q} \mathbf{Z}_{i,t}^{(m)}\right), \quad
\bar{\mathbf{C}}_k = \mathrm{norm}\!\left(\frac{1}{T_q}\sum_{t=1}^{T_q} \mathbf{C}_{k,t}\right)
\end{equation}
\begin{equation}
q_{i,k}^{(m)} = \frac{\exp \left( \bar{\mathbf{Z}}_i^{(m)} \bar{\mathbf{C}}_k^\top / \tau \right)}{\sum_{j=1}^{K_c} \exp \left( \bar{\mathbf{Z}}_i^{(m)} \bar{\mathbf{C}}_j^\top / \tau \right)}
\end{equation}
where $\tau$ is a temperature hyperparameter.

To capture the holistic patient state, we aggregate the assignments across all available modalities to obtain a patient-level consensus distribution:
\begin{equation}
\bar{\mathbf{q}}_i = \frac{1}{|\mathcal{S}_i|} \sum_{m \in \mathcal{S}_i} \mathbf{q}_i^{(m)}
\end{equation}
This consensus $\bar{\mathbf{q}}_i$ serves as a robust, cross-modal anchor that summarizes the patient's comprehensive profile based on available evidence.

\subsubsection{Dynamic Imputation and Refinement}
For any missing modality $m \notin \mathcal{S}_i$, we dynamically impute its token sequence using a mixture of the shared prototypes driven by the consensus distribution. To construct a complete, uniform representation set for patient $i$ across all modalities $m \in \mathcal{M}$, we define the pre-refinement tokens $\mathbf{U}_i^{(m)}$ as:
\begin{equation}
\mathbf{U}_i^{(m)} = 
\begin{cases} 
\mathbf{Z}_i^{(m)}, & \text{if } m \in \mathcal{S}_i \\
\sum_{k=1}^{K_c} \bar{q}_{i,k} \mathbf{C}_k, & \text{if } m \notin \mathcal{S}_i 
\end{cases}
\end{equation}

Finally, to smooth the semantic transition between the originally extracted features and the prototype-imputed features, the unified tokens $\mathbf{U}_i^{(m)}$ are passed through a modality-specific refinement module to generate the final aligned tokens:
\begin{equation}
\tilde{\mathbf{Z}}_i^{(m)} = \text{Transformer}_m\left( \mathbf{U}_i^{(m)} \right) \in \mathbb{R}^{T_q \times D}
\end{equation}
Notably, this imputation and refinement process operates in the latent representation space rather than attempting to reconstruct raw signals, thereby mitigating hallucination risks and ensuring structural alignment across multimodal latent space.

\subsection{Modality-Aware MoE Fusion Backbone}
To capture complex cross-modal interactions while efficiently increasing model capacity, we pass the aligned tokens into a shared backbone. We concatenate the modality tokens in a fixed order to form a fused sequence for patient $i$:
\begin{equation}
\mathbf{X}_i = [\tilde{\mathbf{Z}}^{(\text{I})}_i ; \tilde{\mathbf{Z}}^{(\text{R})}_i ; \tilde{\mathbf{Z}}^{(\text{T})}_i] \in \mathbb{R}^{3T_q \times D}
\end{equation}
A Transformer backbone processes $\mathbf{X}_i$ to produce contextualized tokens. To balance modality-specific specialization and holistic cross-modal context exchange, the backbone alternates between Vanilla Transformer blocks and Sparse Mixture-of-Experts (MoE) blocks. 

Specifically, in the MoE layers, tokens are routed dynamically to top-$k$ experts via a modality-aware gate:
\begin{equation}
\mathbf{r} = W_g(\mathbf{x} + \mathbf{b}_m)
\end{equation}
where $\mathbf{b}_m$ is a learned modality index embedding injected into the gating input to condition the routing on the source modality. We select the top-$k$ experts using $\text{softmax}(\mathbf{r})$ and compute the output as a weighted sum of the selected expert Feed-Forward Networks (FFNs). We further apply a load-balancing regularizer $\mathcal{L}_{\text{router}}$ to encourage uniform expert utilization and prevent routing collapse.

\subsection{Missing-Aware Self-Supervised Pretraining}
Our pretraining paradigm optimizes three complementary objectives: (i) cross-modal alignment before fusion, (ii) post-fusion consistency under structured missingness augmentation, and (iii) MoE routing regularization.

\subsubsection{Pre-Fusion Inter-Modality Alignment}
Before modality fusion, we explicitly align the latent spaces of the available modalities. For each modality $m$, we pool the tokens and apply a projection head $g_m$:
\begin{equation}
\mathbf{v}_i^{(m)} = \text{norm}(g_m(\text{pool}(\tilde{\mathbf{Z}}_i^{(m)}))) \in \mathbb{R}^{D_d}
\end{equation}
where $D_d=256$. Let $a_{i,m} \in \{0, 1\}$ indicate whether modality $m$ is naturally observed for patient $i$. We compute the pairwise InfoNCE loss strictly over the valid cohort where both modalities are present, denoted by the set $\Omega_{m,n} = \{i : a_{i,m} = a_{i,n} = 1\}$:
\begin{equation}
\mathcal{L}_{\text{align}} = \frac{1}{|\mathcal{P}|} \sum_{(m,n) \in \mathcal{P}} \text{InfoNCE}\left(\{\mathbf{v}_i^{(m)}\}_{i \in \Omega_{m,n}}, \{\mathbf{v}_i^{(n)}\}_{i \in \Omega_{m,n}}\right)
\end{equation}
where $\mathcal{P=\{(I,R), (I, T), (R, T)\}}$ represents all valid modality pairs. This masked formulation naturally leverages the union of partially observed cohorts without introducing imputation noise.

\subsubsection{Stochastic Augmentation}
To improve robustness against incomplete multimodal inputs, we propose a Dirichlet-driven stochastic augmentation module to generate two augmented views. 
For each view, we apply sample-wise modality dropout ($p_{mod}$) and token-level dropout ($p_{tok}$) to the observed modalities, yielding a binary keep mask $\mathbf{M}^{(m)}$ that is constrained to preserve at least one valid token per view.

Instead of naive zero-masking, dropped elements are imputed using semantically plausible prototype mixtures. 
Specifically, we first compute the soft assignment $\mathbf{q}_i^{(m)}$ over the shared prototype bank $\mathcal{C}$, and sparsify this distribution by retaining the Top-$K_s$ probabilities with re-normalizing, yielding a valid probability simplex $\hat{\mathbf{q}}_i^{(m)}$.
We then parameterize a Dirichlet distribution using this sparsified distribution scaled by a concentration factor $\alpha$. 
Dirichlet provides simplex-constrained weights centered at $\hat{\mathbf{q}}_i^{(m)}$, with $\alpha$ controlling diversity. 
By sampling continuous mixture weights $\tilde{\mathbf{p}} \sim \text{Dirichlet}(\alpha \cdot \hat{\mathbf{q}}_i^{(m)})$, the augmented tokens for modality $m$ are synthesized as:
\begin{equation}
\tilde{\mathbf{Z}}_i'^{(m)} = \mathbf{M}_i^{(m)} \odot \tilde{\mathbf{Z}}_i^{(m)} + (1 - \mathbf{M}_i^{(m)}) \odot \left( \sum_{k=1}^{K_c} \tilde{p}_{i,k} \mathbf{C}_k \right)
\end{equation}

Overall, this augmentation produces diverse imputed views for contrastive learning while remaining anchored to patient-specific prototype evidence, reducing the risk of implausible imputations under structured missingness.

\subsubsection{Post-Fusion Consistency and Overall Objective}
Both augmented views are then passed through the shared MoE fusion backbone $\mathcal{F}$, producing contextualized tokens $\mathbf{O}_i^{(s)} = \mathcal{F}(\mathbf{X}_i^{(s)}) 
\in \mathbb{R}^{3T_q \times D}$. 
To mitigate the direct contribution of synthetically filled or dropped tokens to the final patient representation, we apply a reliability-weighted pooling scheme. 
Specifically, we define a binary mask $w_{i,t}^{(s)} \in \{0,1\}$ that equals 1 only if token $t$ belongs to a naturally observed modality and was retained during augmentation, and compute:
\begin{equation}
\mathrm{MaskedPool}(\mathbf{O}_i^{(s)}) 
= \frac{\textstyle\sum_{t} w_{i,t}^{(s)} \cdot 
\mathbf{O}_{i,t}^{(s)}}
       {\textstyle\sum_{t} w_{i,t}^{(s)}}
\end{equation}
The pooled features are then mapped via a fusion projection head $g_f$:
\begin{equation}
\mathbf{h}_i^{(s)} = \mathrm{norm}(g_f(\mathrm{MaskedPool}
(\mathbf{O}_i^{(s)}))) \in \mathbb{R}^{D_d}, 
\quad s \in \{1, 2\}
\end{equation}
We enforce representation invariance via a symmetric inter-view InfoNCE loss:
\begin{equation}
\mathcal{L}_{\mathrm{fusion}} = \mathrm{InfoNCE}\left(
\{\mathbf{h}_i^{(1)}\}_i, \{\mathbf{h}_i^{(2)}\}_i\right)
\end{equation}

The overall self-supervised pretraining objective is formulated as a weighted sum:
\begin{equation}
\mathcal{L}_{\text{total}} = \lambda \mathcal{L}_{\text{align}} + (1-\lambda) \mathcal{L}_{\text{fusion}} + \lambda_{\text{router}} \mathcal{L}_{\text{router}}
\end{equation}
where $\lambda$ and $\lambda_{\text{router}}$ are hyperparameters balancing the loss.

\section{Experiments}

\subsection{Dataset and Preprocessing}
\begin{table}[t]
\centering
\caption{Per-cancer cohort statistics for downstream evaluation using tri-modal complete cases.
$N_{\mathrm{tri}}$ denotes the tri-modal cohort size.
OS entries are reported as \textit{event/cens}.
3yOS and 3yRec entries are reported as \textit{pos/neg} after excluding patients censored before 3 years; for 3yRec, excluded cases are those censored before 3 years without a recorded event.}
\label{tab:downstream_cohort}
\setlength{\tabcolsep}{6pt}
\renewcommand{\arraystretch}{1.05}
\begin{tabular}{lcccc}
\specialrule{1.1pt}{0pt}{0pt}
Cancer & $N_{\mathrm{tri}}$ & OS & 3yOS & 3yRec \\
\specialrule{1.1pt}{0pt}{0pt}
UCEC & 499 & 80 / 419  & 62 / 203  & 99 / 176 \\
LUAD & 436 & 157 / 279 & 118 / 114 & 159 / 86 \\
LGG  & 434 & 95 / 339  & 64 / 126  & 121 / 96 \\
BRCA & 855 & 116 / 739 & 56 / 357  & 69 / 340 \\
BLCA & 346 & 161 / 185 & 147 / 74  & 145 / 61 \\
\specialrule{1.1pt}{0pt}{0pt}
\end{tabular}
\end{table}
We curate a TCGA cohort spanning 32 cancer types with up to three modalities per patient from the TCGA repository\footnote{\url{https://portal.gdc.cancer.gov/}}: histopathology whole-slide images (WSI; \textbf{Image}), bulk RNA sequencing (\textbf{RNA}), and pathology reports (\textbf{Text}). Modalities are aligned by TCGA identifiers, yielding 10,439 patients with at least one available modality. Among them, 7,675 patients have complete tri-modal data (Image+RNA+Text), while 2,764 patients are missing one or two modalities. We use the full pan-cancer cohort for self-supervised pretraining \emph{without} accessing any downstream labels.
For downstream fine-tuning and evaluation, we focus on five cancer types with complete tri-modal data (UCEC, LUAD, LGG, BRCA, and BLCA). Cohort statistics are summarized in Table~\ref{tab:downstream_cohort}. 

\subsubsection{Image} We extract patch-level features using a Vision Transformer (ViT) initialized with Marugoto pre-trained weights~\cite{Angeloni2025-uh,Saldanha2023-uf}. Each WSI is tiled into non-overlapping $224\times224$ patches at a fixed magnification. Background/low-information patches are removed using an entropy-based criterion. The remaining patch features are aggregated into a fixed-length sequence of 128 tokens using per-WSI mini-batch $k$-means (128 cluster centroids); zero-padding is applied when fewer than 128 patches are available.

\subsubsection{RNA} We encode bulk RNA-seq profiles using BulkRNABert initialized from published weights~\cite{Gelard2024-tx}. To reduce computation, we use a fixed 2,048-dimensional slice of the produced embeddings for downstream modeling.

\subsubsection{Text} Pathology reports were downloaded from TCGA as PDF files and converted to text via AWS OCR, followed by standard cleaning. The text was tokenized and encoded using a BERT- base Transformer text encoder (BioClinicalBERT~\cite{Alsentzer2019-rv}; 768-d), and sequences were truncated/padded to 200 tokens for minibatch training.

\subsection{Tasks and Evaluation Metrics}
We evaluate models on three clinically relevant endpoints using TCGA annotations.
\textbf{(1) Overall survival (OS) time-to-event prediction.}
Each patient has an observed follow-up time $t_i$ (months) and a censoring indicator $c_i\in\{0,1\}$, where $c_i=1$ denotes right censoring and $c_i=0$ denotes an observed event (death). We adopt a discrete-time survival formulation by discretizing time into $K_{\text{time}}$ intervals and optimize a censoring-aware negative log-likelihood (NLL) loss. Performance is reported using the concordance index (C-index), computed from a scalar risk score derived from the predicted survival distribution.
\textbf{(2) 3-year survival classification.}
We formulate survival beyond 3 years as a binary task with label $y_i^{(\text{sur})}\in\{0,1\}$. Patients censored before 3 years are excluded to ensure unambiguous labels. The model is trained with binary cross-entropy (BCE), and we report AUROC.
\textbf{(3) 3-year recurrence classification.}
Based on TCGA progression-free interval (PFI) annotations, we define $y_i^{(\text{rec})}\in\{0,1\}$ indicating whether recurrence occurs within 3 years among patients. 

\subsection{Baselines and Implementation Details}
\subsubsection{Baselines}
We compare PRIME against unimodal and multimodal baselines under identical downstream training protocols. 
To ensure controlled comparisons, all models take the same \emph{precomputed modality embeddings} as inputs and use the same prediction head design and optimization settings unless otherwise specified. 
This design isolates the contribution of multimodal fusion and pretraining while avoiding confounding effects from differing modality encoders.

\begin{table*}[!t]
\centering
\caption{Task-1 overall survival performance (C-index) on five cancer cohorts. Values are mean$\pm$std over 5-fold CV within each cohort; Avg denotes the mean$\pm$std across cohorts. Best and second-best results are highlighted in \textbf{bold} and \uline{underline}.}
\label{tab:task1_os_cindex}
\footnotesize

\begin{tabular}{lccc ccccc c}
\toprule
\multirow{2}{*}{Methods} & \multicolumn{3}{c}{Modalities} & \multirow{2}{*}{UCEC} & \multirow{2}{*}{LUAD} & \multirow{2}{*}{LGG} & \multirow{2}{*}{BRCA} & \multirow{2}{*}{BLCA} & \multirow{2}{*}{Avg} \\
\cmidrule(lr){2-4}
& Img & RNA & Text & & & & & & \\
\midrule

Img-only   & \checkmark &      &      & 0.668$\pm$0.083 & 0.599$\pm$0.056 & 0.670$\pm$0.100 & 0.615$\pm$0.087 & 0.547$\pm$0.060 & 0.620$\pm$0.052 \\
RNA-only   &       & \checkmark &      & 0.598$\pm$0.094 & 0.510$\pm$0.057 & \uline{0.764$\pm$0.056} & 0.498$\pm$0.065 & 0.526$\pm$0.043 & 0.579$\pm$0.110 \\
Text-only  &       &       & \checkmark & 0.676$\pm$0.066 & 0.553$\pm$0.028 & 0.703$\pm$0.097 & 0.556$\pm$0.030 & 0.532$\pm$0.025 & 0.604$\pm$0.079 \\
ABMIL\cite{Ilse2018-aq} & \checkmark &      &      & 0.637$\pm$0.049 & 0.604$\pm$0.066 & 0.692$\pm$0.056 & 0.617$\pm$0.092 & 0.537$\pm$0.035 & 0.617$\pm$0.056 \\
SNN\cite{Klambauer2017-sq}   &     & \checkmark &      & 0.571$\pm$0.070 & 0.520$\pm$0.065 & 0.719$\pm$0.073 & 0.466$\pm$0.103 & 0.536$\pm$0.053 & 0.563$\pm$0.096 \\
\midrule

Early        & \checkmark & \checkmark & \checkmark & 0.692$\pm$0.045 & 0.557$\pm$0.075 & 0.736$\pm$0.060 & 0.593$\pm$0.050 & 0.557$\pm$0.033 & 0.627$\pm$0.082 \\
Late         & \checkmark & \checkmark & \checkmark & 0.686$\pm$0.068 & 0.575$\pm$0.088 & 0.709$\pm$0.048 & 0.594$\pm$0.024 & 0.528$\pm$0.022 & 0.618$\pm$0.076 \\
CrossAttn   & \checkmark & \checkmark & \checkmark & 0.677$\pm$0.058 & 0.559$\pm$0.097 & 0.688$\pm$0.088 & 0.572$\pm$0.047 & 0.556$\pm$0.037 & 0.610$\pm$0.066 \\
TensorFusion\cite{Zadeh2017-dq} & \checkmark & \checkmark & \checkmark & 0.686$\pm$0.078 & 0.521$\pm$0.048 & 0.734$\pm$0.056 & 0.604$\pm$0.057 & 0.552$\pm$0.049 & 0.620$\pm$0.089 \\
MAGGate\cite{Rahman2020-sm}      & \checkmark & \checkmark & \checkmark & \uline{0.698$\pm$0.049} & 0.553$\pm$0.079 & 0.713$\pm$0.053 & 0.616$\pm$0.035 & 0.538$\pm$0.034 & 0.624$\pm$0.080 \\
MulT\cite{Tsai2019-py}         & \checkmark & \checkmark & \checkmark & 0.677$\pm$0.031 & 0.562$\pm$0.079 & 0.718$\pm$0.086 & 0.576$\pm$0.056 & 0.564$\pm$0.030 & 0.619$\pm$0.073 \\
MCAT\cite{Chen2021-rx}         & \checkmark & \checkmark & & 0.649$\pm$0.041 & 0.589$\pm$0.042 & 0.727$\pm$0.055 & 0.501$\pm$0.077 & 0.533$\pm$0.026 & 0.600$\pm$0.091 \\
Porpoise\cite{Chen2022-pw}     & \checkmark & \checkmark & & 0.663$\pm$0.088 & \textbf{0.631$\pm$0.067} & 0.633$\pm$0.059 & 0.603$\pm$0.068 & 0.559$\pm$0.025 & 0.618$\pm$0.039 \\
PathOmics\cite{Ding2023-ij}    & \checkmark & \checkmark & & 0.609$\pm$0.131 & \uline{0.609$\pm$0.042} & 0.644$\pm$0.070 & 0.572$\pm$0.069 & 0.563$\pm$0.023 & 0.599$\pm$0.033 \\
Song\cite{SongUnknown-db}         & \checkmark & \checkmark & \checkmark & \textbf{0.699$\pm$0.061} & 0.514$\pm$0.054 & 0.684$\pm$0.062 & 0.621$\pm$0.056 & 0.557$\pm$0.068 & 0.615$\pm$0.080 \\
\midrule

Scratch (FT) & \checkmark & \checkmark & \checkmark & 0.684$\pm$0.039 & 0.574$\pm$0.056 & 0.716$\pm$0.064 & 0.573$\pm$0.028 & \uline{0.565$\pm$0.038} & 0.623$\pm$0.072 \\
Scratch (LP)   & \checkmark & \checkmark & \checkmark & 0.684$\pm$0.071 & 0.599$\pm$0.056 & 0.611$\pm$0.085 & 0.560$\pm$0.041 & 0.523$\pm$0.033 & 0.595$\pm$0.060 \\
\midrule

Pretrained (FT) & \checkmark & \checkmark & \checkmark & 0.692$\pm$0.048 & 0.590$\pm$0.051 & 0.730$\pm$0.079 & \textbf{0.643$\pm$0.034} & 0.549$\pm$0.055 & \uline{0.641$\pm$0.073} \\
Pretrained (LP)   & \checkmark & \checkmark & \checkmark & \textbf{0.699$\pm$0.055} & 0.578$\pm$0.051 & \textbf{0.780$\pm$0.050} & \uline{0.622$\pm$0.066} & \textbf{0.584$\pm$0.041} & \textbf{0.653$\pm$0.086} \\
\bottomrule
\end{tabular}
\end{table*}

\textbf{Single-modality baselines.}
We evaluate unimodal baselines that use only one modality at a time. Specifically, \emph{Image-only}, \emph{RNA-only}, and \emph{Text-only} are implemented as single-modality variants of our model, sharing the same backbone, prediction head, and training protocol while restricting the input to the corresponding modality. We additionally include ABMIL~\cite{Ilse2018-aq} for images and SNN~\cite{Klambauer2017-sq} for RNA.

\textbf{Multimodal fusion baselines.}
We include representative fusion strategies: \emph{Early fusion} (feature concatenation followed by an encoder), \emph{Late fusion} (decision-level combination), and \emph{Cross-attention} (cross-modal attention). We further compare with established fusion architectures, including \emph{TensorFusion}~\cite{Zadeh2017-dq}, \emph{MAGGate}~\cite{Rahman2020-sm}, and \emph{MulT}~\cite{Tsai2019-py}, as well as multimodal prognosis models including \emph{MCAT}~\cite{Chen2021-rx}, \emph{Porpoise}~\cite{Chen2022-pw}, \emph{PathOmics}~\cite{Ding2023-ij}, and Song's method~\cite{SongUnknown-db}. For a controlled comparison, Early/Late/Cross-attention/TensorFusion/MAGGate/MulT share the same backbone and differ only in the fusion mechanism. For Task-2/3 (binary classification), baselines originally proposed for survival prediction are adapted by replacing the survival-specific head with a binary classification head while keeping backbone and fusion modules unchanged.

\textbf{Proposed method variants.}
To isolate the effect of pretraining and adaptation strategy, we evaluate our model (i) \emph{from scratch} (random initialization) and (ii) \emph{with self-supervised pretraining}. For each initialization, we report two adaptation modes: \emph{full fine-tuning, FT} (updating the backbone and task head) and \emph{linear probing, LP} (freezing the backbone and training only the task head).

\subsubsection{Implementation details}
We implement all experiments in PyTorch and use identical data splits and task heads across methods for fair comparison.
Pretraining is performed on the unlabeled pan-cancer TCGA cohort (32 cancer types) using AdamW~\cite{Loshchilov2019-on} with learning rate $1\times10^{-5}$, weight decay $0.1$, batch size $64$, and $200$ epochs. We split samples within each cancer type into 80\%/20\% train/validation and select the checkpoint with the lowest validation loss for downstream initialization. The pretraining stage does not access any downstream labels (OS/mortality/recurrence outcomes).
Downstream experiments are conducted on five cancer types (UCEC, LUAD, LGG, BRCA, BLCA) with 5-fold cross-validation, performed within each cancer type. In each fold, patients are split into train/validation/test with a 70/10/20 ratio. We select the best checkpoint on the validation set and report its performance on the held-out test set. Downstream optimization uses AdamW with learning rate $5\times10^{-4}$ (FT) or $1\times10^{-4}$ (LP), batch size $16$, and $50$ epochs. All results are reported as mean$\pm$std across folds and macro-averaged across the five cohorts. Experiments are run on NVIDIA A100-SXM4-40GB GPUs.

\subsection{Results and Analysis}
\subsubsection{Full-modality Performance}
Table~\ref{tab:task1_os_cindex} and \ref{tab:task23_avg_summary} summarize results across three tasks in the full-modality setting. Across all tasks, self-supervised pretraining consistently improves performance over training from scratch, and \emph{Pretrained+LP} achieves the best macro-average results on Task-1 (C-index 0.653) and Task-2 (AUROC 0.689), while \emph{Pretrained+FT} leads on Task-3 (AUROC 0.637). Both pretrained variants outperform all multimodal fusion baselines on the macro-average, indicating that pretraining yields transferable multimodal representations that can be effectively adapted with a lightweight task head.

\begin{table}[!t]
\centering
\caption{Average AUROC across five cancer cohorts for Task-2 (3-year mortality) and Task-3 (3-year recurrence). Values are mean$\pm$std across cohorts. \textbf{Best} and \uline{second-best} methods are highlighted within each task.}
\label{tab:task23_avg_summary}
\footnotesize
\begin{tabular}{lcc}
\toprule
Methods & Task-2 Avg & Task-3 Avg \\
\midrule
Img-only & 0.640$\pm$0.070 & 0.572$\pm$0.087 \\
RNA-only & 0.587$\pm$0.099 & 0.561$\pm$0.117 \\
Text-only & 0.619$\pm$0.070 & 0.608$\pm$0.071 \\
ABMIL\cite{Ilse2018-aq} & 0.638$\pm$0.064 & 0.591$\pm$0.080 \\
SNN\cite{Klambauer2017-sq} & 0.594$\pm$0.110 & 0.568$\pm$0.107 \\
\midrule
Early & 0.655$\pm$0.082 & 0.609$\pm$0.060 \\
Late & 0.646$\pm$0.082 & 0.604$\pm$0.074 \\
CrossAttn & 0.661$\pm$0.063 & 0.628$\pm$0.078 \\
TensorFusion\cite{Zadeh2017-dq} & 0.645$\pm$0.064 & 0.587$\pm$0.045 \\
MAGGate\cite{Rahman2020-sm} & 0.639$\pm$0.074 & 0.598$\pm$0.088 \\
MulT\cite{Tsai2019-py} & 0.650$\pm$0.082 & 0.611$\pm$0.073 \\
MCAT\cite{Chen2021-rx} & 0.626$\pm$0.105 & 0.579$\pm$0.091 \\
Porpoise\cite{Chen2022-pw} & 0.639$\pm$0.072 & 0.579$\pm$0.103 \\
PathOmics\cite{Ding2023-ij} & 0.621$\pm$0.092 & 0.593$\pm$0.074 \\
Song\cite{SongUnknown-db} & \uline{0.672$\pm$0.108} & 0.626$\pm$0.092 \\
\midrule
Scratch (FT) & 0.656$\pm$0.064 & 0.610$\pm$0.068 \\
Scratch (LP) & 0.636$\pm$0.072 & 0.595$\pm$0.081 \\
\midrule
Pretrained (FT) & 0.669$\pm$0.077 & \textbf{0.637$\pm$0.056} \\
Pretrained (LP) & \textbf{0.689$\pm$0.110} & \uline{0.629$\pm$0.098} \\
\bottomrule
\end{tabular}
\end{table}

\begin{figure*}[!t]
\centerline{\includegraphics[width=\textwidth]{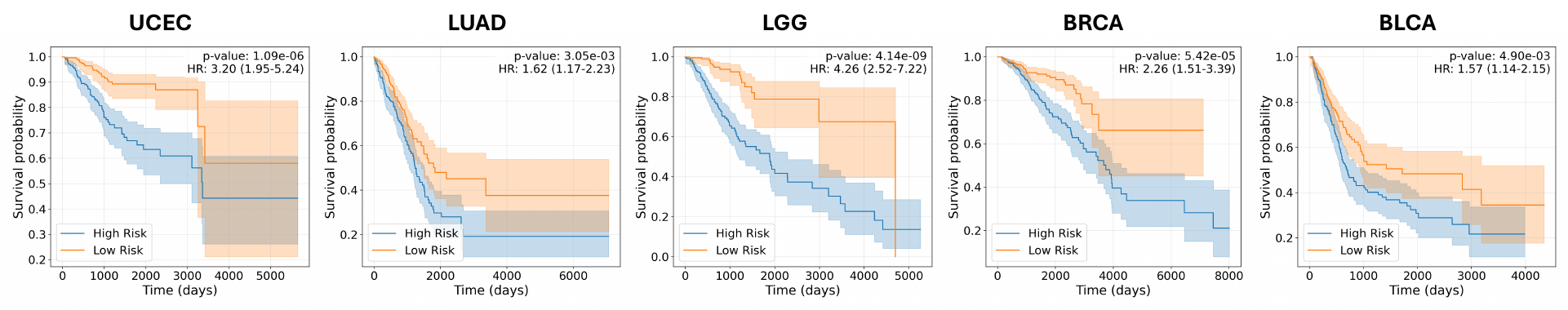}}
\caption{Kaplan--Meier survival curves based on predicted risk scores for five TCGA cohorts. Each panel reports the log-rank $p$-value and the hazard ratio (HR) from a univariate Cox model comparing high-risk vs.\ low-risk groups.}
\label{fig_km}
\end{figure*}

\begin{table*}[t]
\centering
\caption{Robustness to missing modalities on three downstream tasks. Models are trained on full tri-modal data, with missingness introduced only at test time. Full denotes all modalities; LI/LR/LT and OI/OR/OT denote leave-one-out and only-one settings for Image/RNA/Text, respectively. Values are mean$\pm$std over 5-fold CV.}
\label{tab:robustness_missing}
\footnotesize
\setlength{\tabcolsep}{3.2pt}
\renewcommand{\arraystretch}{1.08}

\begin{tabular}{c|c|ccccccc}
\toprule
Task & Method & Full & LI & LR & LT & OI & OR & OT \\
\midrule
\multirow{3}{*}{Task-1} 
& Scratch (FT)         & 0.623$\pm$0.072 & 0.608$\pm$0.069 & 0.631$\pm$0.068 & 0.607$\pm$0.051 & 0.607$\pm$0.050 & 0.540$\pm$0.085 & 0.606$\pm$0.066 \\
& Pretrained (LP) & 0.653$\pm$0.086 & 0.612$\pm$0.086 & 0.641$\pm$0.086 & 0.619$\pm$0.090 & 0.603$\pm$0.058 & 0.564$\pm$0.084 & 0.611$\pm$0.077 \\
& Pretrained (LP+Missing) & 0.653$\pm$0.089 & 0.623$\pm$0.085 & 0.639$\pm$0.085 & 0.636$\pm$0.087 & 0.611$\pm$0.055 & 0.571$\pm$0.063 & 0.610$\pm$0.089 \\
\midrule
\multirow{3}{*}{Task-2} 
& Scratch (FT)         & 0.656$\pm$0.064 & 0.611$\pm$0.052 & 0.666$\pm$0.064 & 0.621$\pm$0.073 & 0.620$\pm$0.072 & 0.486$\pm$0.051 & 0.611$\pm$0.052 \\
& Pretrained (LP) & 0.689$\pm$0.110 & 0.670$\pm$0.118 & 0.671$\pm$0.097 & 0.665$\pm$0.088 & 0.638$\pm$0.077 & 0.618$\pm$0.112 & 0.641$\pm$0.101 \\
& Pretrained (LP+Missing) & 0.679$\pm$0.117 & 0.679$\pm$0.118 & 0.670$\pm$0.104 & 0.676$\pm$0.089 & 0.642$\pm$0.077 & 0.627$\pm$0.118 & 0.646$\pm$0.107 \\
\midrule
\multirow{3}{*}{Task-3} 
& Scratch (FT)         & 0.610$\pm$0.068 & 0.587$\pm$0.076 & 0.624$\pm$0.052 & 0.584$\pm$0.071 & 0.585$\pm$0.071 & 0.524$\pm$0.097 & 0.587$\pm$0.075 \\
& Pretrained (LP) & 0.629$\pm$0.098 & 0.612$\pm$0.078 & 0.611$\pm$0.090 & 0.620$\pm$0.114 & 0.594$\pm$0.097 & 0.577$\pm$0.108 & 0.597$\pm$0.064 \\
& Pretrained (LP+Missing) & 0.622$\pm$0.100 & 0.624$\pm$0.084 & 0.632$\pm$0.096 & 0.612$\pm$0.115 & 0.609$\pm$0.095 & 0.574$\pm$0.109 & 0.626$\pm$0.070 \\
\bottomrule
\end{tabular}
\end{table*}

Notably, linear probing matches or surpasses full fine-tuning after pretraining, whereas the opposite holds when training from scratch (e.g., Task-1: Scratch+LP 0.595 vs.\ Scratch+FT 0.623; Pretrained+LP 0.653 vs.\ Pretrained+FT 0.641). This suggests that pretraining produces sufficiently structured representations for parameter-efficient adaptation. While cohort-level variation exists (e.g., Porpoise leads on LUAD for Task-1), our pretrained variants provide the most consistently strong performance across cancer types and endpoints.

\subsubsection{Risk Stratification via Kaplan--Meier Analysis}
Beyond rank-based evaluation (C-index), we assess whether the predicted risk scores can stratify patients into clinically distinct survival groups. 
For each cancer cohort, we pool held-out test predictions across the 5-fold cross-validation splits and apply a median threshold on the predicted risk score to define \emph{high-risk} and \emph{low-risk} groups. 
We plot Kaplan--Meier survival curves and report the log-rank test $p$-value. 
To quantify effect size, we fit a univariate Cox proportional hazards model with a binary covariate indicating the high-risk group and report the hazard ratio (HR) with 95\% confidence intervals.

As shown in Fig.~\ref{fig_km}, the predicted risk scores yield clear separation between the two survival curves in all five cohorts, with statistically significant log-rank tests and HR$>1$ for the high-risk group. 
For example, on BRCA, the high-risk group exhibits significantly worse survival with HR$=2.26$ (95\% CI $1.51$--$3.39$) and a significant log-rank test ($p=5.42\times10^{-5}$), demonstrating substantial effect size beyond statistical significance. 
These results indicate that the model learns clinically meaningful risk stratification signals rather than only improving a ranking metric. 

\subsubsection{Robustness to Missing Modalities}
We evaluate robustness by introducing missingness at test time while training and validating on full tri-modal data (Table~\ref{tab:robustness_missing}). 
This controlled setting isolates the model's ability to operate under incomplete inputs at inference and avoids confounding effects from missingness during supervised training. 
Two settings are considered: \textbf{leave-one-out} (LI/LR/LT), where one modality is removed, and \textbf{only-one} (OI/OR/OT), where a single modality is available.
We compare Scratch (FT), Pretrained (LP), and a variant Pretrained (LP+Missing) that applies modality dropout during downstream training.

Across all three tasks and missingness patterns, pretrained representations substantially outperform scratch, with the largest gains in the only-one setting where Scratch (FT) degrades most severely. For instance, on Task-2 OR (RNA-only), Scratch achieves 0.486 while Pretrained (LP) reaches 0.618, a gap of over 0.13. In general, RNA-only (OR) causes the largest degradation, suggesting that RNA features benefit most from cross-modal pretraining.

Introducing modality dropout at downstream training (LP+Missing) further improves robustness under missingness, sometimes at a small cost on the full-modality score. For example, on Task-3, LR improves from 0.611 to 0.632 and OT from 0.597 to 0.626, while full-modality performance decreases only marginally (0.629 to 0.622). This trade-off is generally favorable in clinical settings where modality completeness cannot be guaranteed.



\subsubsection{Label Efficient Downstream Transfer}
\begin{figure}[!t]
\centerline{\includegraphics[width=\columnwidth]{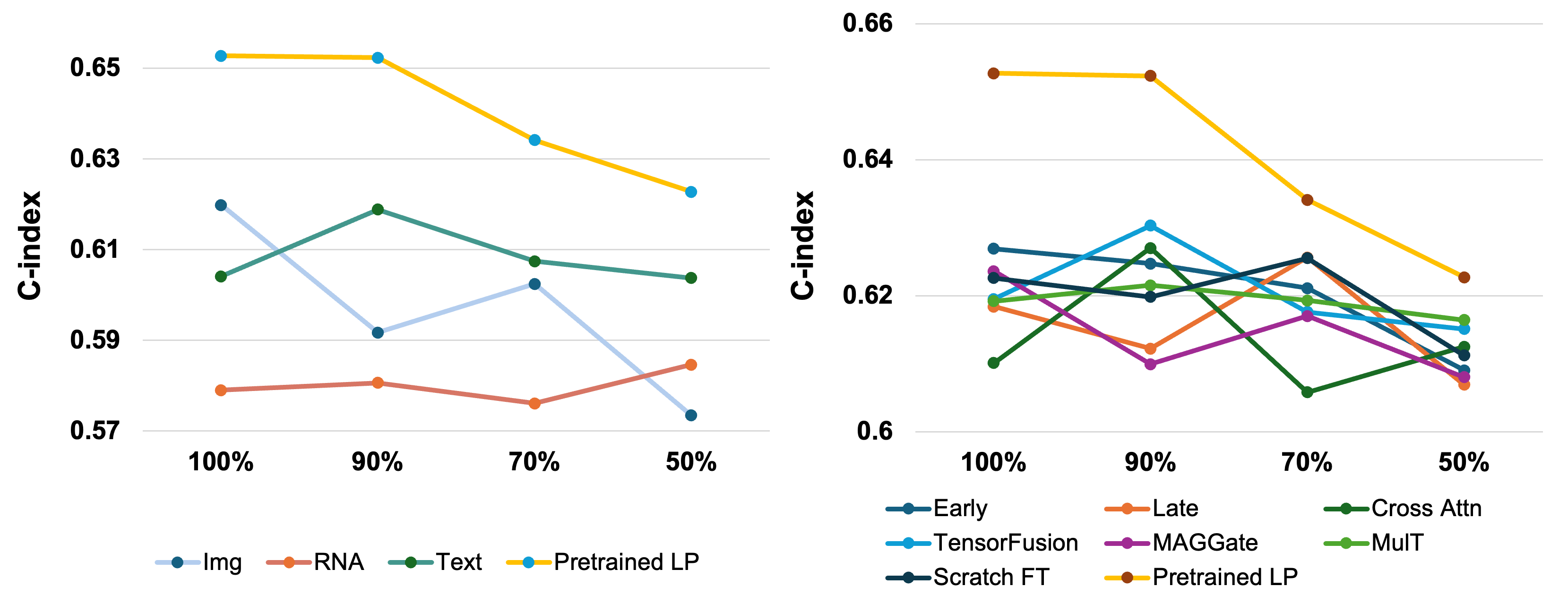}}
\caption{Label-efficient downstream transfer performance on 
Task-1 (overall survival; C-index). The x-axis denotes the 
fraction of labeled \emph{training} samples retained from the 
full training split. Left: unimodal baselines. Right: 
multimodal fusion baselines.}
\label{fig2}
\end{figure}

We further conduct a label efficient transfer study on Task-1 by downsampling the labeled training set while keeping the validation/test splits unchanged. Specifically, for each fold we randomly retain \{100\%, 90\%, 70\%, 50\%\} of the labeled training samples with the same sampling indices across methods, and report the corresponding macro-averaged C-index. Fig.~\ref{fig2} summarizes the results.

As shown in Fig.~\ref{fig2} (left), our method consistently outperforms unimodal baselines (\emph{Image-only}, \emph{RNA-only}, and \emph{Text-only}) across all label budgets and exhibits a noticeably smaller performance degradation as the training set shrinks. This indicates that the pretrained multimodal representation reduces sample complexity and remains effective even when supervision is limited.
Fig.~\ref{fig2} (right) further compares our method with representative fusion baselines and a scratch-trained FT baseline. Across all sampling ratios, our method achieves the best C-index and degrades more gracefully under reduced supervision, suggesting that self-supervised pretraining yields transferable cross-modal features that can be reliably adapted in the low-data regime.
Overall, these results demonstrate the advantage of our pretraining-and-adaptation design for practical settings where labeled outcomes are limited.

\subsubsection{Parameter-Efficiency}
Pretraining also improves parameter-efficient adaptation. While linear probing is weaker than full fine-tuning when trained from scratch, \emph{Pretrained (LP)} achieves strong downstream performance, attaining the best macro-average results on Task-1 (0.653) and Task-2 (0.689), and remaining competitive on Task-3 (0.629) compared with \emph{Pretrained (FT)} (0.637). These results suggest that the learned multimodal representations are highly transferable and can be effectively adapted using lightweight task-specific heads, offering a favorable accuracy--efficiency trade-off.

\begin{table*}[!t]
\centering
\caption{Ablation on key components of our pretraining framework. All variants use the same downstream protocol (pretrained + LP) and are evaluated on the five-cohort average (mean$\pm$std over 5-fold CV). \textbf{Full method} uses missing-aware pretraining, prototypes, and the combined objective $\mathcal{L}=\lambda \mathcal{L}_{\mathrm{align}}+(1-\lambda) \mathcal{L}_{\mathrm{fusion}}$. ``w/o $\mathcal{L}_{\mathrm{align}}$'' corresponds to $\lambda=0$, and ``w/o $\mathcal{L}_{\mathrm{fusion}}$'' corresponds to $\lambda=1$.}
\label{tab:ablation_components}
\footnotesize
\setlength{\tabcolsep}{4.0pt}
\renewcommand{\arraystretch}{1.08}

\begin{tabular}{lcccc|ccc}
\toprule
\multirow{2}{*}{Variant} 
& \multicolumn{4}{c|}{Components / Objective}
& \multicolumn{3}{c}{Downstream performance} \\
\cmidrule(lr){2-5}\cmidrule(lr){6-8}
& \makecell{Pretrained Data \\ (missing+full modality)} & Prototypes & $L_{\mathrm{align}}$ & $L_{\mathrm{fusion}}$
& T1 (C-index) & T2 (AUROC) & T3 (AUROC) \\
\midrule

w/o prototypes (token=0) 
& $\checkmark$ & $\times$ & $\checkmark$ & $\checkmark$
& \uline{0.646$\pm$0.082} & \uline{0.669$\pm$0.123} & 0.595$\pm$0.108 \\

w/o missing modality
& $\times$ & $\checkmark$ & $\checkmark$ & $\checkmark$
& 0.640$\pm$0.074 & 0.656$\pm$0.116 & \textbf{0.629$\pm$0.087} \\

\midrule
\textbf{Full method (Pretrained LP)} 
& $\checkmark$ & $\checkmark$ & $\checkmark$ & $\checkmark$
& \textbf{0.653$\pm$0.086} & \textbf{0.689$\pm$0.110} & \uline{0.629$\pm$0.098} \\
\midrule

w/o $L_{\mathrm{align}}$ ($\lambda=0$) 
& $\checkmark$ & $\checkmark$ & $\times$ & $\checkmark$
& 0.614$\pm$0.072 & 0.625$\pm$0.087 & 0.626$\pm$0.065 \\

w/o $L_{\mathrm{fusion}}$ ($\lambda=1$) 
& $\checkmark$ & $\checkmark$ & $\checkmark$ & $\times$
& 0.589$\pm$0.068 & 0.624$\pm$0.129 & 0.609$\pm$0.093 \\

\bottomrule
\end{tabular}
\end{table*}
\subsubsection{Ablation Studies}
We conduct ablation studies to quantify the contribution of key components in our pretraining framework. Table~\ref{tab:ablation_components} summarizes the performance under a controlled setting where all variants share the same downstream protocol (\emph{pretrained+LP}). We include the MoE router load-balancing regularizer $\mathcal{L}_{\text{router}}$ in all variants with the same value; thus it is held fixed and not ablated here. The \textbf{full method} combines missing-aware pretraining, a learnable prototype bank, and the joint objective $L=\lambda L_{\mathrm{align}}+(1-\lambda)L_{\mathrm{fusion}}$.

\textbf{Impact of prototypes and missing-aware pretraining.}
Replacing prototype tokens with zeros (w/o prototypes) leads to consistent degradations, particularly on Task-2 and Task-3 (T1: $0.653\!\rightarrow\!0.646$, T2: $0.689\!\rightarrow\!0.669$, T3: $0.629\!\rightarrow\!0.595$), indicating that learnable prototypes provide informative shared tokens for imputing missing modalities and strengthening multimodal transfer. Disabling missing-aware pretraining (pretraining only on full-modality samples) also reduces Task-1 and Task-2 (T1: $0.653\!\rightarrow\!0.640$, T2: $0.689\!\rightarrow\!0.656$). In contrast, Task-3 shows a negligible difference in the mean (both $0.629$), suggesting that this endpoint may be less sensitive to exposure to missing-modality patterns during pretraining, or that the effect is masked by the relatively large cross-fold variance.

\textbf{Role of the pretraining objective.}
Ablating either loss term yields a clear drop from the full objective, demonstrating that the two terms provide complementary supervision. Using only $L_{\mathrm{fusion}}$ (w/o $L_{\mathrm{align}}$, $\lambda=0$) harms Task-1 and Task-2 (T1: $0.653\!\rightarrow\!0.614$, T2: $0.689\!\rightarrow\!0.625$), highlighting the importance of explicit cross-modal alignment for learning transferable representations. Using only $L_{\mathrm{align}}$ (w/o $L_{\mathrm{fusion}}$, $\lambda=1$) further reduces Task-1 and notably impacts Task-3 (T1: $0.653\!\rightarrow\!0.589$, T3: $0.629\!\rightarrow\!0.609$), suggesting that fusion-level consistency provides additional training signals beyond alignment and is important for exploiting cross-modal interactions.

Overall, these ablations demonstrate that (i) the learnable prototype bank and missing-aware pretraining both contribute to robust transfer under modality incompleteness, and (ii) combining $L_{\mathrm{align}}$ and $L_{\mathrm{fusion}}$ is necessary to obtain strong and balanced performance across tasks.

\subsubsection{Sensitivity Analysis}
\begin{figure}[!t]
\centerline{\includegraphics[width=\columnwidth]{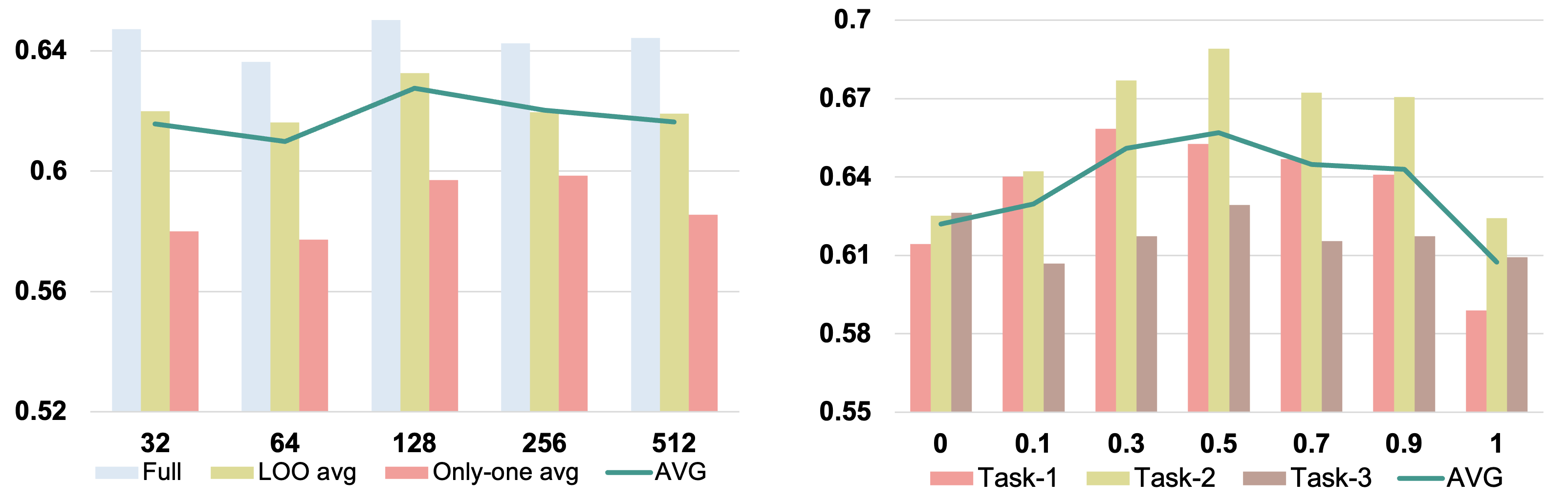}}
\caption{Sensitivity of the model to prototype bank size $K_c$ (left) and loss weight coefficient $\lambda$ (right).}
\label{fig3}
\end{figure}

We analyze sensitivity to two key hyperparameters: the learnable prototype bank size $K_c$ and the loss weighting coefficient $\lambda$. All results follow the same evaluation protocol as above. When varying one hyperparameter, we keep the other fixed (we use $\lambda=0.5$ for the $K_c$ sweep and $K_c=128$ for the $\lambda$ sweep).

\textbf{Prototype bank size $K_c$.}
Fig.~\ref{fig3} (left) reports an overall score in task 1 under three input conditions: \emph{Full} (all modalities), \emph{LOO avg} (leave-one-out missing-modality average), and \emph{Only-one avg} (single-modality average), together with their mean (\emph{AVG}). 
Performance remains stable across a wide range of $K_c$, while a moderate bank size yields the best trade-off. Increasing $K_c$ from 32/64 to 128 improves robustness under missingness (LOO avg peaks $0.633$ at $K_c=128$; Only-one avg improves to $0.597$--$0.598$ at $K_c=128$--$256$), and the overall mean reaches its maximum at $K_c=128$ (AVG $0.628$). Further enlarging the bank to $K_c=512$ provides diminishing returns and slightly degrades the single-modality case, likely due to redundancy and more difficult prototype retrieval. We therefore use $K_c=128$ by default.

\textbf{Loss weight $\lambda$.}
Fig.~\ref{fig3} (right) shows that balancing $L_{\mathrm{align}}$ and $L_{\mathrm{fusion}}$ is important. 
The best overall performance is achieved $\lambda=0.5$ (AVG $0.657$), where Task-2 also peaks (AUROC $0.689$) and Task-3 is strongest (AUROC $0.629$). Moving away from this balance degrades performance, and the extremes are clearly suboptimal: using only fusion ($\lambda=0$) reduces the average (AVG $0.622$), while using only alignment ($\lambda=1$) leads to the largest drop (AVG $0.607$), with a notable decrease on Task-1 (C-index $0.589$). 
These trends suggest that $L_{\mathrm{align}}$ and $L_{\mathrm{fusion}}$ provide complementary supervision: $L_{\mathrm{align}}$ encourages modality-invariant representations for transfer, whereas $L_{\mathrm{fusion}}$ regularizes multimodal aggregation to better exploit cross-modal interactions, especially under missingness.

Overall, the model is not overly sensitive to hyperparameter choices, but both analyses favor moderate settings ($K_c=128$ and $\lambda=0.5$) that yield consistently strong and balanced performance across tasks and missing-modality conditions.

\subsubsection{Discussion and Limitations}
Our results indicate that large-scale, label-free multimodal pretraining can provide a practical foundation for cancer outcome modeling from heterogeneous clinical inputs. The prototype memory bank and missing-aware design consistently improve robustness when one modality is unavailable at inference, which is common in pathology workflows where genomics or complete reports may be missing. The strong performance of linear probing further suggests that the pretrained backbone captures transferable cross-modal structure, enabling accurate adaptation with minimal task-specific parameters. Ablation and sensitivity analyses also support that the alignment and fusion losses provide complementary supervision and that moderate hyperparameter choices lead to stable performance.

This study has several limitations. First, we primarily evaluate missingness by removing modalities at test time while training on complete tri-modal data; this isolates inference-time robustness but does not fully reflect settings where supervised training data are also incomplete. Second, we standardize all methods to use precomputed modality embeddings to enable controlled comparisons of fusion and pretraining strategies, which may understate the potential benefits of end-to-end encoder fine-tuning. 
Third, recent multimodal pretraining frameworks such as mSTAR~\cite{Xu2025-ib}, POMP~\cite{Wang2025-hw}, and MICE~\cite{Zhou2025-co} are not included as direct baselines because they differ from PRIME in pretraining design: MICE incorporates supervised survival objectives, mSTAR targets image-only inference via knowledge distillation with cancer-type supervision, and POMP requires fully paired data without addressing missing modalities. All three are also tightly coupled with specific encoders, precluding controlled comparison under our unified protocol with shared frozen embeddings.
Finally, additional validation on external cohorts is needed to assess robustness under domain shift and support deployment in safety-critical decision support.

\section{Conclusion}
\label{sec:Conclusion}
We propose a large-scale multimodal pretraining framework for cancer prognosis on TCGA histopathology WSIs, RNA-seq, and pathology reports. By combining missing-modality-aware pretraining, a learnable prototype memory bank, and a joint objective, our method learns transferable representations that support robust prediction under heterogeneous and incomplete clinical inputs. Across five cancer cohorts and three clinically relevant endpoints, self-supervised pretraining consistently improves downstream performance, while linear probing remains highly competitive with full fine-tuning using fewer trainable parameters. Robustness experiments, ablations, and sensitivity studies further validate the contributions of each component. 
Overall, our results highlight the promise of missing-aware multimodal foundation model pretraining as a practical basis for reliable prediction and decision support in safety-critical settings where modality incompleteness is common.
Our code will publicly available at 
\url{https://github.com/yukkai/PRIME}.

\bibliographystyle{IEEEtran}
\bibliography{paperpile}

\end{document}